\documentclass[11pt]{article}
\usepackage{multirow}
\usepackage[preprint]{acl}

\usepackage{times}
\usepackage{latexsym}
\usepackage{amsmath}
\usepackage{booktabs}
\usepackage{subcaption}
\usepackage{graphicx}
\usepackage{tcolorbox}
\usepackage{xcolor}
\usepackage[T1]{fontenc}

\usepackage[utf8]{inputenc}
\usepackage[utf8]{inputenc}
\DeclareUnicodeCharacter{2194}{\ensuremath{\leftrightarrow}}
\usepackage{microtype}

\usepackage{inconsolata}

\usepackage{graphicx}

\usepackage{fancyhdr}

\fancypagestyle{logostyle}{
    \fancyhf{} 
    \lhead{\includegraphics[height=3em]{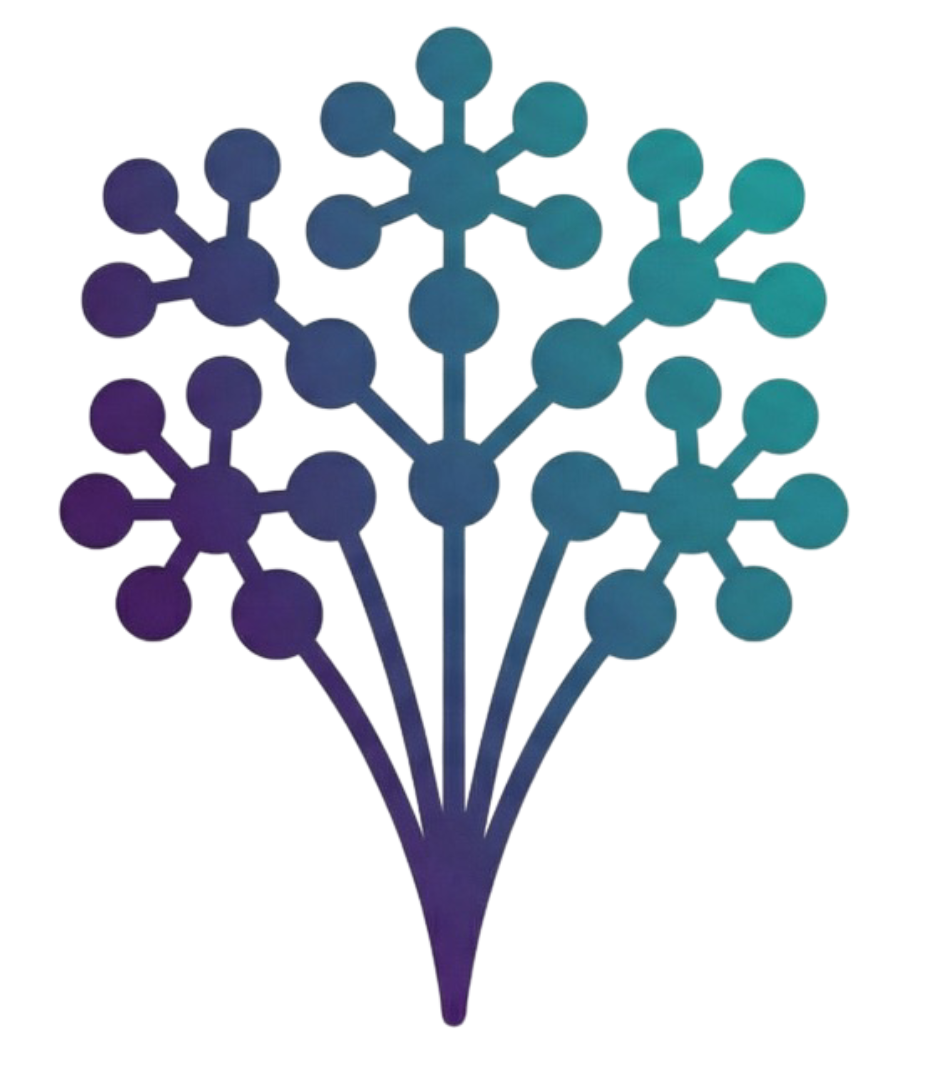}} 
     
}

\title{
    \begin{minipage}[c]{0.10\textwidth} 
        \includegraphics[width=\linewidth]{Figures/bouquet_logo.png}
    \end{minipage}%
    \hfill 
    \begin{minipage}[c]{0.85\textwidth}
        \raggedright 
        \bfseries    
        \Large       
        A Hybrid Protocol for Large-Scale Semantic Dataset Generation in Low-Resource Languages: The Turkish Semantic Relations Corpus
    \end{minipage}
    
    \vspace{-0.1cm} 
}

\author{
  Ebubekir Tosun \\
  \texttt{ebubekirsoftware@gmail.com} \\
  \And
  Mehmet Emin Buldur \\
  \texttt{mehmeteminbuldur@outlook.com} \\
  \AND  
  Özay Ezerceli \\
  \texttt{ezerceli804@gmail.com} \\
  \And
  Mahmoud ElHussieni \\
  \texttt{mahmoud.elhussieni@outlook.com} 
}

\begin{document}
\maketitle
\begin{abstract}
We present a hybrid methodology for generating large-scale semantic relationship datasets in low-resource languages, demonstrated through a comprehensive Turkish semantic relations corpus. Our approach integrates three phases: (1) FastText embeddings with Agglomerative Clustering to identify semantic clusters, (2) Gemini 2.5-Flash for automated semantic relationship classification, and (3) integration with curated dictionary sources. The resulting dataset comprises 843,000 unique Turkish semantic pairs across three relationship types (synonyms, antonyms, co-hyponyms) representing a 10× scale increase over existing resources at minimal cost (\$65). We validate the dataset through two downstream tasks: an embedding model achieving 90\% top-1 retrieval accuracy and a classification model attaining 90\% F1-macro. Our scalable protocol addresses critical data scarcity in Turkish NLP and demonstrates applicability to other low-resource languages. We publicly release the dataset and models.
\end{abstract}

\section{Introduction}

Turkish, despite having a population of over 88 million speakers, lacks comprehensive semantic relationship datasets comparable to those available for English and other resource-rich languages \cite{miller1995wordnet,navigli2012babelnet}. This scarcity impedes the development of semantic parsing systems, word sense disambiguation tools, and semantic similarity models for Turkish NLP applications. While significant efforts have been made to construct Turkish resources through manual annotation and dictionary mining \cite{bakay-etal-2021-turkish,ehsani2018kenet}, the limited scale and domain coverage of existing resources present a bottleneck for modern neural NLP systems that require millions of training examples \cite{devlin2019bert}.

The challenge of building semantic resources for morphologically rich languages like Turkish is compounded by several factors. First, the agglutinative nature of Turkish means that a single root can generate hundreds of valid word forms through productive suffixation, requiring substantially larger vocabularies to achieve equivalent coverage compared to analytic languages. Second, existing resources such as Turkish Tree Bank \cite{ehsani2018kenet} and KeNet \cite{bakay-etal-2021-turkish} rely primarily on translation-based projections from English WordNet or limited manual curation, inheriting biases and leaving domain-specific terminology (particularly in legal, medical, and technical domains) largely uncovered. Third, the cost of manual annotation at scale is prohibitive for academic research budgets.

We propose a hybrid protocol that targets both computational efficiency and linguistic quality. The method proceeds in three phases.

\paragraph{Context Preparation} We start from an expert-curated lexicon of 77,000 terms and expand it to 110,000 entries using Named Entity Recognition–based augmentation. We then compute FastText embeddings and apply Agglomerative Clustering, yielding 13,000 semantic clusters that act as contextual structure for downstream relationship classification.
\paragraph{LLM-Based Semantic Enrichment} In the second phase, we employ Gemini 2.5-Flash to automatically induce and label diverse semantic relationships within clusters. The model exploits both cluster-level context and its multilingual knowledge to generate high-quality relationship labels at an overall cost of about \$65.
\paragraph{Dictionary Integration} Finally, we incorporate an external Turkish Synonym Dictionary comprising roughly 20,000 entries. After applying strict filters that retain only high-precision cases with at most two synonym candidates, we obtain 16,000 validated synonym pairs.

This hybrid design couples large-scale automation with explicit quality control and yields 843,000 unique semantic pairs, while dictionary-based validation helps preserve semantic coherence across the induced relations.

We make three primary contributions:

\begin{enumerate}
    \item A \textbf{scalable hybrid methodology} combining embedding-based clustering with LLM enrichment that is generalizable to other low-resource languages with minimal language-specific adaptation.

    \item The \textbf{Turkish Semantic Relations Corpus}, comprising 843,000 annotated semantic pairs across three relationship types (synonym, antonym, co-hyponym) in JSONL format, representing the largest native Turkish semantic resource to date.

    \item \textbf{Validation through downstream models}: an embedding model achieving 90\% top-1 retrieval accuracy for synonym pairs and a classification model achieving 90\% F1-macro, demonstrating the dataset's utility for training production-quality semantic systems.
\end{enumerate}

\section{Related Work}
\subsection{Semantic Resources for Low-Resource Languages}

The construction of large-scale semantic resources has historically been dominated by manual curation efforts. Princeton WordNet \cite{miller1995wordnet} established the paradigm of organizing lexical knowledge through synsets connected by semantic relations, inspiring similar projects across dozens of languages through the Global WordNet Association \cite{bond2012survey}. However, the labor-intensive nature of WordNet construction has limited the scale and coverage achievable for under-resourced languages.

For Turkish specifically, several resources have been developed with varying approaches. Turkish Tree Bank \cite{ehsani2018kenet} was constructed through statical machine translation of English WordNet synsets, inheriting both the conceptual organization and potential cultural biases of the source resource. KeNet \cite{bakay-etal-2021-turkish} took a more native approach, building from Turkish dictionaries and corpora, but remains limited in coverage with approximately 80,000 synsets. These resources, while valuable, cover primarily general vocabulary and lack depth in specialized domains.

\subsection{LLM-Augmented Dataset Generation}

The emergence of large language models has opened new possibilities for dataset generation at scale. Recent work has demonstrated that LLMs can effectively generate training data for various NLP tasks, including semantic similarity \cite{schick2021generating}, natural language inference \cite{wang2021want}, and question answering \cite{alberti2019synthetic}. The key insight is that LLMs, having been trained on massive multilingual corpora, encode substantial world knowledge that can be extracted through careful prompting.

Several studies have specifically examined LLM-generated semantic relationships. \citet{chen2023llm} showed that GPT-4 can generate high-quality synonym and antonym pairs with accuracy comparable to crowdsourced annotations when provided with appropriate context. \citet{kumar2023data} demonstrated that data augmentation through LLM paraphrasing significantly improves downstream task performance for low-resource settings. Our work extends this line of research by combining LLM generation with clustering-based context provision and dictionary validation.

\subsection{Embedding-Based Clustering for Semantic Organization}

Distributional semantics provides a foundation for organizing terms into semantically coherent groups. Word embeddings such as Word2Vec \cite{mikolov2013distributed}, GloVe \cite{pennington2014glove}, and FastText \cite{Bojanowski_Grave_Joulin_Mikolov_2016} capture semantic similarity through co-occurrence patterns, enabling unsupervised discovery of related terms. FastText is crucial for morphologically rich languages, particularly because it provides meaningful representations for rare word forms and novel compound words thanks to its sub-word modeling capabilities.

Agglomerative clustering on embedding spaces has been successfully applied to various semantic tasks. \citet{lin2015clustering} used hierarchical clustering to induce word sense inventories, while \citet{haghighi2008learning} employed similar techniques for coreference resolution. Our approach uses clustering not as a final output but as contextual input for LLM-based relationship classification, providing semantic scaffolding that improves generation quality.

\subsection{Antonym-Synonym Distinction in Vector Spaces}

A core difficulty in distributional semantics is that antonyms and synonyms frequently receive highly similar vector representations because they tend to appear in near-identical contexts \cite{mohammad2013computing}. The expressions “the water is hot” and “the water is cold”, for instance, share the same syntactic frame, which leads standard embedding models to place “hot” and “cold” in close proximity in the vector space despite their clear semantic opposition \cite{mohammad2013computing}.

Counter-fitting \cite{mrksic2016counter} mitigates this issue by post‑processing pretrained embeddings so that antonym pairs are pushed apart while synonym pairs are drawn together, but this strategy depends on precompiled lexical resources to specify the corresponding constraints \cite{mrksic2016counter}. In contrast, our three-way classification scheme (synonym/antonym/co-hyponym) directly encodes this distinction via supervised learning, allowing the resulting classifier to filter out antonyms from synonym candidates without the need for manually crafted constraint sets.

\section{Methodology}

\begin{figure*}[t]
\centering
\includegraphics[width=0.99\textwidth]{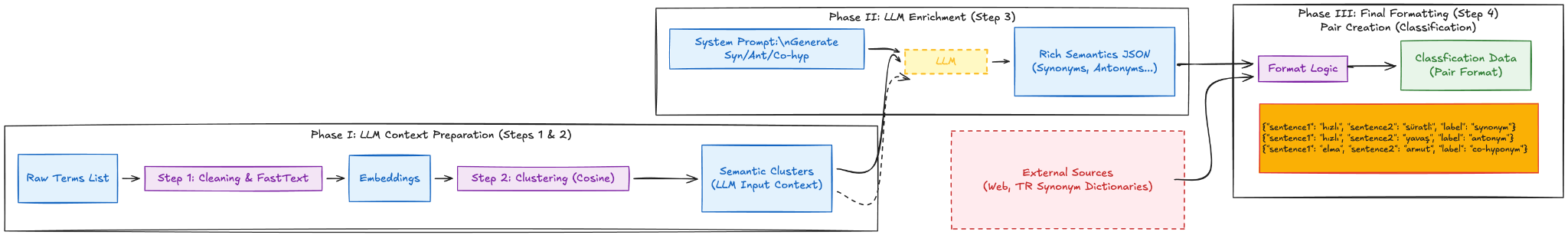}
\caption{Overview of the three-phase hybrid protocol for semantic dataset generation. Phase I establishes semantic structure through FastText embeddings (110K terms) and agglomerative clustering (13K clusters with distance threshold 0.4). Phase II employs Gemini 2.5-Flash to classify intra-cluster relationships into three types: synonyms (strict equivalence), antonyms (semantic opposition), and co-hyponyms (thematic relatedness). Phase III integrates 16K high-precision dictionary pairs and outputs 843K annotated pairs in JSONL format.}
\label{fig:pipeline}
\end{figure*}

Our dataset generation pipeline comprises three sequential phases, each addressing specific challenges in creating high-quality semantic relationship data at scale. Figure~\ref{fig:pipeline} illustrates the complete workflow.

\subsection{Phase I: Context Preparation}

\paragraph{Initial Term Collection}
The foundation of our dataset is an expert-curated list of 77,000 legal and domain-specific concept-terms assembled by legal specialists over several years. This list emphasizes terminology from Turkish legal codes, court decisions, and regulatory documents, providing strong coverage of formal register vocabulary.

To increase domain breadth, we augmented this list with unique concept-terms extracted from Named Entity Recognition (NER) datasets accumulated during prior information extraction work. Our NER system was trained to identify legal concepts, technical terms, and domain-specific entities in Turkish text, providing a complementary source of specialized vocabulary. This expansion yielded a final list of approximately 110,000 unique terms.

\paragraph{Embedding Generation}
We generated vector representations for all terms using the pre-trained Facebook FastText Turkish model (cc\_tr\_300), which provides 300-dimensional embeddings trained on Common Crawl and Wikipedia data. FastText’s subword modeling is particularly valuable for Turkish, as it enables meaningful representations for morphological variants sharing common stems, compound terms composed of known subwords, and domain-specific terminology that may be absent from the original training vocabulary.

For multi-word expressions, we compute the embedding as the mean of constituent word vectors, providing reasonable representations for phrases and compound terms.

\paragraph{Semantic Clustering}
We applied Agglomerative Clustering with cosine distance metric on the FastText embeddings:

\begin{equation}
d(u, v) = 1 - \frac{u \cdot v}{\|u\| \|v\|}
\end{equation}

A distance threshold of 0.4 was selected intentionally to group terms based on broad thematic relevance rather than strict synonymy. This relatively lenient threshold ensures that true synonyms are highly likely to be co-clustered, while antonyms—which often share similar distributional contexts—may also appear within the same cluster, alongside co-hyponyms organized under a common hypernym.

This process yielded approximately 13,000 semantic clusters ranging in size from 2 to 50+ terms. These clusters serve as contextual input for the LLM enrichment phase, providing semantic scaffolding that improves relationship classification accuracy.

\subsection{Phase II: LLM-Based Semantic Enrichment}

\paragraph{Model Selection}
Following optimization experiments balancing performance and computational cost, we selected \textbf{Gemini 2.5-Flash} \cite{comanici2025gemini25} as the generator model. This choice was motivated by its strong multilingual capabilities, including Turkish, cost-effective API pricing (\$0.075 per 1M input tokens), consistent generation of structured outputs, and a sufficiently large context window (up to 1M tokens) that enables efficient batch processing. The total cost for processing all 13{,}000 clusters was approximately \$65.

The total cost for processing all 13,000 clusters was approximately \$65.

\paragraph{Prompt Design}
We designed a comprehensive system prompt that instructs the model to analyze each cluster and classify semantic relationships between terms into three categories: synonyms, defined as terms with identical or near-identical meanings that are fully substitutable in context; antonyms, defined as terms exhibiting exact semantic opposition; and co-hyponyms, defined as terms that share a common hypernym and thematic category while remaining distinct in their specific meanings.

The prompt includes explicit categorization rules with illustrative examples, strict golden rules prohibiting uncertain classifications, and requirements for structured JSON output formatting. Key design choices enforce strict synonymy by accepting only exact synonyms while assigning near-synonyms to the co-hyponym category, treat abbreviations and their expansions as valid synonym pairs (e.g., ``VUK'' $\leftrightarrow$ ``Vergi Usul Kanunu''), encourage the creative generation of semantically equivalent multi-word expressions, and instruct the model to augment each cluster with additional semantic relationships drawn from its internal knowledge.

\paragraph{Batch Processing}
Clusters were processed in batches via the Gemini API using multiprocessing for parallelization. For each cluster, the model outputs structured JSON objects that map each distinct concept to its semantic relationships (i.e., synonyms, antonyms, and co-hyponyms). Post-processing steps include removing self-synonyms (where a term appears in its own synonym list), deduplicating relationship pairs, normalizing Unicode representations, and validating the structural integrity of the generated JSON outputs. This phase produced approximately 827,000 labeled semantic pairs.

\subsection{Phase III: Integration and Final Formatting}

\paragraph{Dictionary Integration}
To augment the synthetically generated data with high-confidence ground truth, we integrated an external Turkish synonym dictionary (T\"{u}rk\c{c}e E\c{s} Anlaml{\i}lar S\"{o}zl\"{u}\u{g}\"{u}) containing approximately 20{,}000 entries. To ensure high precision, we applied strict filtering by retaining only headwords with at most two synonym candidates, excluding entries with ambiguous or context-dependent meanings, and removing any entries that overlapped with LLM-generated data.

This filtering yielded 16,000 high-reliability pairs that serve as validation anchors within the larger synthetic dataset.

\paragraph{Format Standardization}
The final dataset is stored in JSONL format with the following structure:

\begin{verbatim}
{"sentence1": "term_A",
 "sentence2": "term_B",
 "label": "synonym|antonym|co_hyponym"}
\end{verbatim}

This format is directly compatible with standard sentence-pair classification frameworks and enables straightforward conversion to contrastive learning formats.

\section{Dataset Analysis}

\subsection{Distribution Statistics}

Table~\ref{tab:distribution} presents the complete distribution statistics of our corpus. The dataset comprises 842,946 total pairs, with co-hyponyms representing 71.96\% (606,612 pairs), synonyms 17.60\% (148,367 pairs), and antonyms 10.44\% (87,967 pairs). The synthetic LLM-generated pairs constitute 98.10\% of the dataset, while dictionary-derived pairs account for 1.90

\begin{table}[t]
\centering
\small 
\begin{tabular}{lrr}
\toprule
\textbf{Metric} & \textbf{Count} & \textbf{\%} \\ 
\midrule
\multicolumn{3}{l}{\textit{Class Distribution}} \\
\quad Co-hyponym & 606,612 & 71.96\% \\ 
\quad Synonym & 148,367 & 17.60\% \\
\quad Antonym & 87,967 & 10.44\% \\
\midrule
\textbf{Total Pairs} & \textbf{842,946} & 100.00\% \\
\midrule
\multicolumn{3}{l}{\textit{Source Distribution}} \\
\quad Synthetic (LLM) & 826,946 & 98.10\% \\
\quad Dictionary & 16,000 & 1.90\% \\
\midrule
\multicolumn{3}{l}{\textit{Textual Statistics}} \\
\quad Avg. Word Count & \multicolumn{2}{c}{$\sim$2.0} \\
\quad Vocabulary Diversity (TTR) & \multicolumn{2}{c}{0.02} \\
\quad Avg. Token Length & \multicolumn{2}{c}{11.04} \\
\quad Max. Token Length & \multicolumn{2}{c}{37} \\
\bottomrule
\end{tabular}
\caption{Detailed statistics of the Turkish Semantic Relations Corpus. The class distribution reflects natural language patterns where co-hyponyms are more frequent than synonyms or antonyms.}
\label{tab:distribution}
\end{table}

The observed class imbalance, with co-hyponyms accounting for 72\% of all pairs, mirrors the general tendency in natural language for broad semantic relatedness to be far more frequent than strict synonymy or antonymy. This skewed distribution has direct implications for model behavior and therefore motivates the use of weighted loss functions to prevent the classifier from overfitting to the dominant co-hyponym class.

The low Type–Token Ratio (0.02) indicates a highly interconnected pairwise structure in which the same anchor terms participate in multiple, distinct semantic relations. In line with this, tokenization statistics show a maximum input length of 37 tokens (mean 11.04), which empirically supports the choice of 64 as the upper bound on sequence length during classifier training and avoids unnecessary padding overhead.

\subsection{Domain Coverage}

Roughly 4--5\% of the generated instances contain foreign legal terminology (e.g., English or French expressions) that is routinely used in Turkish legal practice, reflecting the inherently international orientation of contemporary legal systems. Within this setting, the dataset spans the following domains:

The dataset covers a wide range of domain-specific vocabulary, including legal terminology (e.g., contract, criminal, administrative, and constitutional law), financial terms spanning banking, insurance, taxation, and corporate finance, technical vocabulary from information technology, engineering, and medicine, as well as administrative language related to government procedures, institutional entities, and regulatory concepts.

\section{Experiments and Results}

We validate the utility of our dataset through two downstream tasks: contrastive embedding learning and relationship classification.

\subsection{Embedding Model}

\subsubsection{Data Preparation}

The embedding training data consists of approximately 55{,}000 unique samples organized as $(\textit{query}, \textit{positive}, \textit{hard\_negatives})$ triplets, where positive examples correspond to terms labeled as true synonyms, and hard negatives consist of terms labeled as antonyms or co-hyponyms.

Including co-hyponyms as hard negatives made the performance of the model worse than the scenario where don't use co-hyponums. So, it is not crucial for forcing the model to distinguish between strict semantic equivalence and broad thematic similarity.

\subsubsection{Model Architecture}

We utilize multilingual-e5-large \cite{wang2024multilinguale5textembeddings} (560M parameters, XLM-RoBERTa architecture) as the backbone encoder in a Siamese configuration. The embedding is computed via mean pooling:

\begin{equation}
\mathbf{u} = \frac{\sum_{i=1}^{L} \mathbf{H}_i \cdot M_i}{\sum_{i=1}^{L} M_i}
\end{equation}

where $\mathbf{H}_i$ represents hidden states and $M_i$ is the attention mask.

\subsubsection{Training Configuration}

We employ Cached Multiple Negatives Ranking Loss (CMNRL) \cite{gao2021scalingdeepcontrastivelearning}, which expands the negative sample set using cached gradients from previous batches:

\begin{equation}
\mathcal{L}_i = -\log \frac{e^{\text{sim}(u_i, v_i)/\tau}}{e^{\text{sim}(u_i, v_i)/\tau} + \sum_{j \in \mathcal{B} \setminus \{i\} \cup \mathcal{C}} e^{\text{sim}(u_i, v_j)/\tau}}
\end{equation}

Training hyperparameters are detailed in Table~\ref{tab:embed_config}.

\begin{table}[t]
\centering
\small
\begin{tabular}{ll}
\toprule
\textbf{Parameter} & \textbf{Value} \\
\midrule
Base Model & multilingual-e5-large \\
Optimizer & AdamW (fused) \\
Learning Rate & $3 \times 10^{-5}$ \\
Scheduler & Cosine with warmup (0.1) \\
Batch Size & 128 \\
Epochs & 8 \\
Temperature ($\tau$) & 0.07 \\
Max Sequence Length & 512 \\
Precision & BF16 \\
Hardware & NVIDIA RTX 3060 (12GB) \\
\bottomrule
\end{tabular}
\caption{Embedding model training configuration.}
\label{tab:embed_config}
\end{table}

\subsubsection{Embedding Model Results}

\begin{figure*}[t] 
    \centering
    \begin{subfigure}[b]{0.48\textwidth} 
        \centering
        \includegraphics[width=\linewidth]{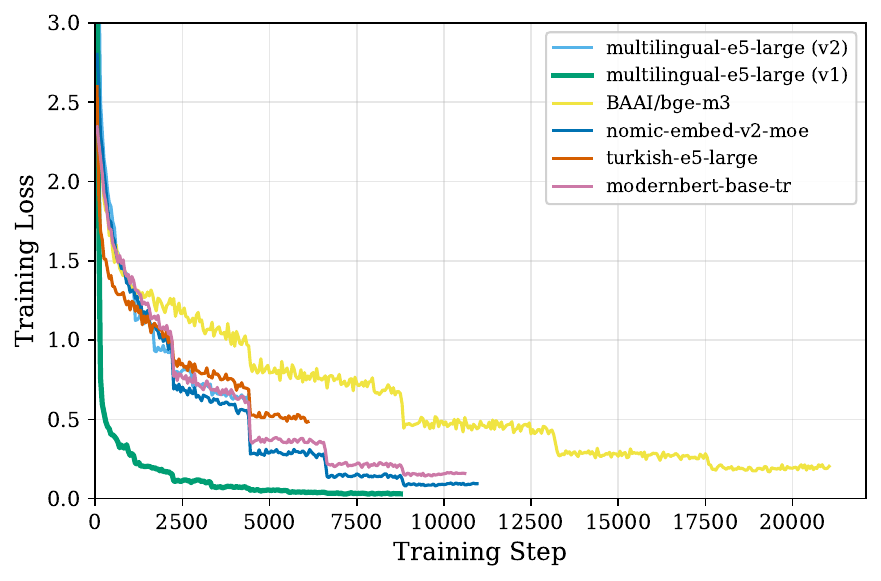}
        \caption{Training Loss}
    \end{subfigure}
    \hfill
    \begin{subfigure}[b]{0.48\textwidth}
        \centering
        \includegraphics[width=\linewidth]{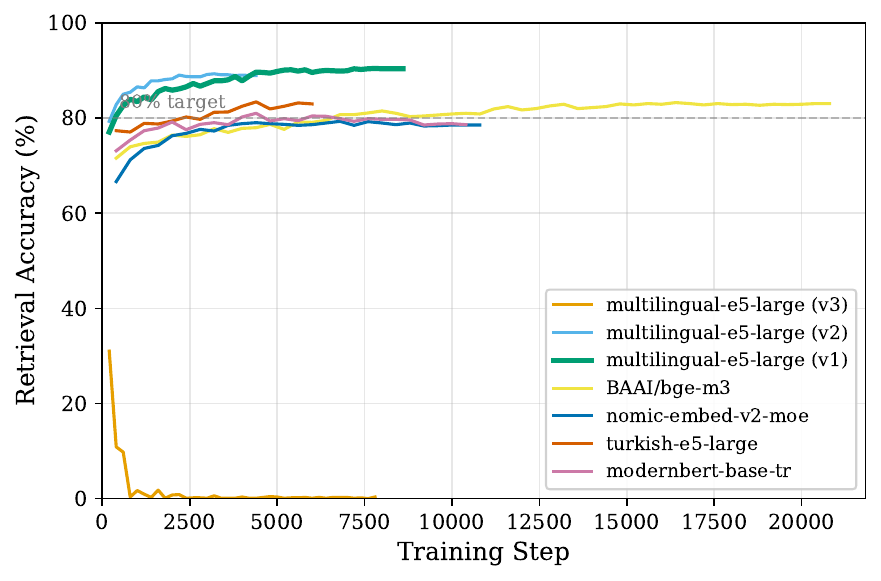}
        \caption{Retrieval Accuracy}
    \end{subfigure}
    \caption{Embedding model training progression across seven model candidates. The \texttt{multilingual-e5-large} (v1) variant achieves best performance with 90\% retrieval accuracy.}

    \label{fig:embed_results_wide}
\end{figure*}

The embedding model achieves \textbf{90\% top-1 retrieval accuracy for synonym pairs} on a held-out test set, where accuracy is measured as the proportion of queries for which the true synonym appears in the top-k retrieved results. Figure~\ref{fig:embed_results_wide} shows the training dynamics.

\subsection{Classification Model}

\subsubsection{Phase 1: Model Candidate Selection}

Six models spanning different architectures and parameter sizes were evaluated:
\paragraph{Model Selection}
We evaluated six transformer-based sentence embedding models: TurkEmbed4STS~\cite{ezerceli2025turkembed}, a Turkish-specific model trained on semantic textual similarity tasks; modernbert-base-tr~\cite{newmindai2025modernbert}, a recent Turkish adaptation of the ModernBERT architecture; mpnet~\cite{allmpnetbasev2}, a widely-used multilingual sentence encoder; bert-base-turkish-128k~\cite{bertbaseturkishcased}, a Turkish BERT variant with extended context; turkish-e5-large~\cite{izdas2025turkish}, a large-scale Turkish embedding model; and multilingual-e5-large~\cite{wang2024multilingual}, a multilingual variant. Performance comparison is presented in Table~\ref{tab:model_selection}. Based on superior F1-macro performance (0.87) and stable training dynamics, turkish-e5-large was selected for subsequent experiments.

\subsubsection{Experimental Framework}

We established a two-phase experimental framework consisting of an initial model selection phase, in which six candidate models were benchmarked under identical experimental conditions, followed by a final training phase that performed optimized training of the selected model on upgraded hardware.

\subsubsection{Phase 1: Model Candidate Selection}

Six models spanning different architectures and parameter sizes were evaluated:

\begin{table}[t]
\centering
\small
\begin{tabular}{lcc}
\toprule
\textbf{Model} & \textbf{Params} & \textbf{F1-Macro} \\
\midrule
TurkEmbed4STS & 305M & 0.82 \\
modernbert-base-tr & 135M & 0.79 \\
mpnet & 278M & 0.83 \\
bert-base-turkish-128k & 184M & 0.81 \\
turkish-e5-large & 560M & \textbf{0.87}\\
multilingual-e5-large & 560M & 0.85 \\
\bottomrule
\end{tabular}
\caption{Phase I model comparison results. All models trained for 5 epochs with identical hyperparameters on NVIDIA RTX 3060.}
\label{tab:model_selection}
\end{table}

The \texttt{turkish-e5-large} model demonstrated superior performance, achieving 0.87 F1-macro with stable convergence.

\subsubsection{Phase 2: Final Optimized Training}

The selected model was retrained with upgraded configuration on NVIDIA L40S hardware with extended context processing capabilities. Table~\ref{tab:final_config} summarizes the complete training setup, including the optimized batch size of 128, learning rate of $3 \times 10^{-5}$, and maximum sequence length of 64 tokens based on our empirical tokenization analysis.
\begin{table}[t]
\centering
\small
\begin{tabular}{ll}
\toprule
\textbf{Parameter} & \textbf{Value} \\
\midrule
Base Model & turkish-e5-large \\
Hardware & NVIDIA L40S \\
Batch Size & 128 \\
Learning Rate & $3 \times 10^{-5}$ \\
Epochs & 5 \\
Max Length & 64 \\
Gradient Clipping & 1.0 \\
Precision & BF16 \\
\bottomrule
\end{tabular}
\caption{Phase 2 final training configuration.}
\label{tab:final_config}
\end{table}

\subsubsection{Classification Model Results}

The final model achieves \textbf{90\% F1-macro} on the held-out test set. Table~\ref{tab:class_results} presents the detailed per-class performance metrics. Notably, the minority synonym class achieves 0.83 F1-score despite representing only 17.60\% of training data, while antonyms (0.92 F1) and co-hyponyms (0.94 F1) demonstrate even stronger performance. The weighted loss function successfully mitigates class imbalance, with the macro-averaged precision and recall reaching 0.88 and 0.92 respectively.

\begin{table}[t]
\centering
\begin{tabular}{lccc}
\toprule
\textbf{Class} & \textbf{Precision} & \textbf{Recall} & \textbf{F1} \\
\midrule
Synonym & 0.76& 0.90& 0.83\\
Antonym & 0.91& 0.93& 0.92\\
Co-hyponym & 0.93 & 0.95 & 0.94 \\
\midrule
\textbf{Macro Avg.} & 0.88& 0.92& \textbf{0.90}\\
\bottomrule
\end{tabular}
\caption{Per-class classification results for the final model.}
\label{tab:class_results}
\end{table}

The weighted loss function successfully addresses class imbalance, with minority classes (synonym, antonym) achieving competitive performance despite comprising only 28\% of training data.

\section{Discussion and Limitations}

\subsection{Strengths of the Hybrid Approach}

Our methodology combines complementary strengths by leveraging FastText-based clustering to provide scalable semantic organization without requiring labeled data, employing LLM-based enrichment to capture nuanced semantic relationships that distance-based metrics alone cannot distinguish, and integrating curated dictionary resources as validation anchors to ensure a reliable baseline level of quality.

The three-way classification (synonym/antonym/co-hyponym) reflects linguistic theory while supporting discriminative models. Co-hyponym classification is particularly valuable, as these relationships represent shared semantic space without synonymy. Distinction crucial for models that must capture both similarity and specificity.

\subsection{Limitations}

\paragraph{Domain Bias}
The dataset is primarily grounded in legal domain vocabulary, which may introduce systematic biases. Models trained on this data may underperform on casual or conversational Turkish.

\paragraph{Synthetic Data Proportion}
Approximately 98\% of the data is synthetically generated via LLM. While our human evaluation shows high quality, LLM-specific biases may propagate to downstream models.

\paragraph{Static Resource}
The dataset represents a snapshot of terminology as of 2025. Legal and technical vocabulary evolves, requiring periodic updates.

\paragraph{Morphological Coverage}
While our terms include various morphological forms, we do not systematically expand across the full paradigm of Turkish suffixation. A term like ``karar'' (decision) may not have all its inflected forms (kararlar{\i}, karar{\i}nda, etc.) represented.

\subsection{Generalizability}

The hybrid protocol is designed for cross-linguistic transfer:

\begin{enumerate}
    \item \textbf{Phase I} requires only FastText embeddings (available for 157 languages) and standard clustering algorithms
    \item \textbf{Phase II} requires an LLM with target language capability (increasingly available through multilingual models)
    \item \textbf{Phase III} requires a dictionary resource (widely available)
\end{enumerate}

We estimate the protocol could be applied to any language with FastText embeddings and basic dictionary resources at comparable cost (\$50--100 for LLM API calls).

\section{Conclusion}

We present a scalable hybrid protocol for generating large-scale semantic relationship datasets in low-resource languages. The Turkish Semantic Relations Corpus comprises 843,000 annotated semantic pairs combining LLM enrichment with dictionary validation, representing the largest native Turkish semantic resource to date. Validation through downstream embedding (90\% retrieval accuracy) and classification (90\% F1-macro) models demonstrates practical utility.

The protocol's independence from extensive manual annotation makes it generalizable to other low-resource languages facing similar data scarcity challenges. We release the dataset and trained models to facilitate research in under-resourced languages.\footnote{Dataset and models will be released upon acceptance.}









\bibliography{custom}

\newpage

\appendix

\section*{Appendices}
\addcontentsline{toc}{section}{Appendices}

\section{Representative Dataset Examples}
\label{app:examples}

Table~\ref{tab:examples} presents representative samples from the Turkish Semantic Relations Corpus across all three relationship types. These examples illustrate the diversity of semantic relationships captured in the dataset, ranging from legal terminology (sözleşme/mukavele for contract), financial oppositions (alıcı/satıcı for buyer/seller), and domain-specific co-hyponyms (hukuk/ceza for civil law/criminal law). English translations are provided in parentheses to facilitate understanding for international readers. The examples demonstrate the dataset's coverage of both common vocabulary and specialized legal/financial terminology.

\begin{table}[hp]
\centering
\footnotesize
\resizebox{\columnwidth}{!}{%
\begin{tabular}{lll}
\toprule
\textbf{Term 1} & \textbf{Term 2} & \textbf{Relation} \\
\midrule
sözleşme & mukavele & Synonym (contract) \\
mahkeme & yargı & Synonym (court/judiciary) \\
alıcı & satıcı & Antonym (buyer/seller) \\
aktif & pasif & Antonym (active/passive) \\
hukuk & ceza & Co-hyponym (civil law/criminal law) \\
banka & sigorta & Co-hyponym (bank/insurance) \\
\bottomrule
\end{tabular}%
}
\caption{Example semantic pairs from the Turkish Semantic Relations Corpus demonstrating the three relationship types: synonyms, antonyms, and co-hyponyms.}
\label{tab:examples}
\end{table}

\section{LLM Semantic Enrichment Prompt Template}
\label{app:llm_prompt}

The complete system prompt used for LLM-based semantic enrichment in Phase II is provided in Figure~\ref{fig:llm_prompt}. This prompt instructs Gemini 2.5-Flash to classify semantic relationships within each cluster into three categories (synonyms, antonyms, co-hyponyms) while following strict categorization rules and output formatting requirements.

\begin{figure*}[tp]
    \centering
    \begin{tcolorbox}[
        colback=gray!10,
        colframe=gray!60,
        boxrule=0.5pt,
        arc=2mm,
        left=5pt, right=5pt,
        top=5pt, bottom=5pt
    ]
    \scriptsize\ttfamily
    You are an expert linguistic system specializing in semantic relationship classification for Turkish language. Your task is to analyze clusters of related terms and classify the semantic relationships between them.\\
    
    \#\#\# RELATIONSHIP TYPES:\\
    
    \textbf{1. SYNONYM (Eş Anlamlı):}\\
    * Terms with identical or nearly identical meanings\\
    * 100\% substitutable in context without meaning change\\
    * Examples: "sözleşme" ↔ "mukavele" (contract), "mahkeme" ↔ "yargı" (court)\\
    * Include abbreviations and their expansions: "VUK" ↔ "Vergi Usul Kanunu"\\
    
    \textbf{2. ANTONYM (Zıt Anlamlı):}\\
    * Terms with exact semantic opposition\\
    * Direct opposites on a semantic scale\\
    * Examples: "alıcı" ↔ "satıcı" (buyer/seller), "aktif" ↔ "pasif" (active/passive)\\
    
    \textbf{3. CO-HYPONYM (Eş Üst Kavram):}\\
    * Terms belonging to the same thematic category\\
    * Share a common hypernym but distinct in specific meaning\\
    * Near-synonyms that are not 100\% substitutable\\
    * Examples: "hukuk" + "ceza" (civil law + criminal law, both are law types)\\
    * Examples: "banka" + "sigorta" (bank + insurance, both are financial institutions)\\
    
    \#\#\# GOLDEN RULES:\\
    
    * STRICT SYNONYMY: Only mark as synonym if 100\% substitutable\\
    * NO UNCERTAIN CLASSIFICATIONS: Skip if relationship is unclear\\
    * AUGMENT FREELY: Add related terms from your knowledge\\
    * STRUCTURED OUTPUT: Return valid JSON only\\
    * NO SELF-SYNONYMS: A term cannot be its own synonym\\
    
    \#\#\# OUTPUT FORMAT:\\
    
    Return a JSON object mapping each term to its relationships:\\
    \{\\
    \quad "term\_1": \{\\
    \quad\quad "synonyms": ["syn\_1", "syn\_2"],\\
    \quad\quad "antonyms": ["ant\_1"],\\
    \quad\quad "co\_hyponyms": ["cohyp\_1", "cohyp\_2"]\\
    \quad \},\\
    \quad ...\\
    \}\\
    
    \#\#\# EXAMPLE:\\
    
    Input cluster: ["mahkeme", "yargı", "adalet"]\\
    Output:\\
    \{\\
    \quad "mahkeme": \{\\
    \quad\quad "synonyms": ["yargı", "mahkeme dairesi"],\\
    \quad\quad "antonyms": [],\\
    \quad\quad "co\_hyponyms": ["adalet", "hukuk sistemi"]\\
    \quad \},\\
    \quad ...\\
    \}
    \end{tcolorbox}
    \caption{System prompt template for LLM-based semantic relationship classification in Phase II. The prompt enforces strict categorization rules and structured JSON output to ensure consistency across all 13,000 processed clusters.}
    \label{fig:llm_prompt}
\end{figure*}

\section{NER-Based Term Augmentation Prompt}
\label{app:ner_prompt}

The system prompt used for Named Entity Recognition-based term augmentation in Phase I (Context Preparation) is shown in Figure~\ref{fig:ner_prompt}. This prompt guides the LLM to extract domain-specific concepts, terms, events, and facts from legal documents, enabling expansion of the initial 77,000-term lexicon to 110,000 entries. The prompt distinguishes between abstract concepts (e.g., "justice", "freedom"), technical terms (e.g., "administrative fine", "copyright"), time-bound events (e.g., "court decision", "contract signing"), and timeless facts (e.g., legal regulations, systemic features).

\begin{figure*}[tp!]
    \centering
    \begin{tcolorbox}[
        colback=gray!10,
        colframe=gray!60,
        boxrule=0.5pt,
        arc=2mm,
        left=5pt, right=5pt,
        top=5pt, bottom=5pt
    ]
    \scriptsize\ttfamily
    You are an expert system in taxonomy and information extraction. Your goal is to detect ALL concepts, terms, events, and facts in the text provided by the user completely, systematically, and consistently, and place them into the correct classes.\\
    
    \#\#\# BASIC DEFINITIONS AND CLASSIFICATION:\\
    
    \textbf{CONCEPT:}\\
    * The abstract and general framework of meaning formed in the mind, covering common characteristics of objects, events, or thoughts\\
    * Mental representations used in making sense of, classifying, and categorizing the world\\
    * Abstract qualities, values, principles, states (e.g., "justice", "freedom", "responsibility", "right", "trust")\\
    * Expressions carrying philosophical, ethical, social, or general scientific meaning\\
    
    \textbf{TERM:}\\
    * Words or groups of words specific to science/profession/art/technical fields, given special meanings in specific contexts\\
    * Special expressions providing precise and unambiguous definition\\
    * Legal, technical, scientific, professional terminology (e.g., "administrative fine", "copyright", "protein synthesis")\\
    * Established institution/board/procedure names and technical idioms\\
    
    \textbf{EVENT:}\\
    * Situations occurring within a specific period, with a specific start and end time and location\\
    * Concrete, observable, singular and unique actions or situations\\
    * Singular situations, decisions, transactions that occurred at a specific time/place\\
    * Action and its object should be evaluated together (e.g., "the court making a decision", "signing of the contract")\\
    
    \textbf{FACT:}\\
    * Facts independent of time, generally valid truths, or existing situations\\
    * Objective, undisputed, accepted realities and principles\\
    * Normative obligations, situations indicating continuity\\
    * Legal regulations, general rules, systemic features\\
    
    \#\#\# ANALYSIS RULES:\\
    
    \textbf{Not to be Accepted:}\\
    * General words frequently used in daily language ("expense", "cost", "time", "area")\\
    * Only proper names (person, institution, place names)\\
    * Conjunctions, pronouns, prepositions, words without meaning\\
    * Specific names of certain legal regulations ("Law X", "Regulation Y")\\
    
    \textbf{Evaluation Criteria:}\\
    * Concept vs Term: Distinction between mental representation (concept) vs linguistic label (term)\\
    * Event vs Fact: Time-dependent singular situation (event) vs timeless general rule (fact)\\
    * Contextual analysis: The same word can fall into different classes in different contexts\\
    * Capture multi-word expressions fully within their natural boundaries\\
    * If possible, evaluate action + object together for EVENT\\
    
    \textbf{Technical Rules:}\\
    * Must be verbatim as it appears in the text, character for character, including punctuation marks\\
    * Do not repeat the same expression (case insensitive)\\
    * Return only expressions explicitly present in the text\\
    * Do not add comments, inferences, or additional explanations\\
    \end{tcolorbox}
    \caption{System prompt template for NER-based term augmentation in Phase I. This prompt enabled extraction of 33,000 additional domain-specific terms from legal corpora, expanding the initial lexicon from 77,000 to 110,000 entries.}
    \label{fig:ner_prompt}
\end{figure*}

\end{document}